\documentclass{article}

\PassOptionsToPackage{numbers,sort}{natbib}

\usepackage[preprint]{nips_2018}

\usepackage[utf8]{inputenc} 
\usepackage[T1]{fontenc}    
\usepackage{hyperref}       
\usepackage{url}            
\usepackage{booktabs}       
\usepackage{amsfonts}       
\usepackage{nicefrac}       
\usepackage{microtype}      
\usepackage{epsfig}
\usepackage{subfigure}
\usepackage{calc}
\usepackage{amssymb}
\usepackage{amstext}
\usepackage{amsmath}
\usepackage{amsthm}
\usepackage{multicol}
\usepackage{pslatex}
\usepackage{ dsfont }
\usepackage{ mathrsfs }

\usepackage[utf8]{inputenc} 
\usepackage[T1]{fontenc} 
\usepackage{palatino} 
\usepackage{multirow}
\usepackage{mathptmx} 
\usepackage{times} 
\usepackage{amsmath} 
\usepackage{amssymb}
\usepackage{algorithm}
\usepackage{url} 

\usepackage{algpseudocode}

\makeatletter
\def\BState{\State\hskip-\ALG@thistlm}
\makeatother

\DeclareMathOperator*{\argmin}{arg\,min}

\title{Deep Quality-Value (DQV) Learning}

\author{
  Matthia Sabatelli\\
  Montefiore Institute\\
  Universit\'e de Li\`ege, Belgium \\
  \texttt{m.sabatelli@uliege.be} \\
  \And 
  Gilles Louppe \\
  Montefiore Institute\\
  Universit\'e de Li\`ege, Belgium \\
  \texttt{g.louppe@uliege.be} \\
  \And 
  Pierre Geurts \\
  Montefiore Institute\\
  Universit\'e de Li\`ege, Belgium \\
  \texttt{p.geurts@uliege.be} \\
  \And 
  Marco A. Wiering \\
  Bernoulli Institute for Mathematics, Computer Science \\ and Artificial Intelligence \\
  University of Groningen, The Netherlands \\
  \texttt{m.a.wiering@rug.nl} \\
}

\begin{document}

\maketitle

\begin{abstract}
We introduce a novel Deep Reinforcement Learning (DRL) algorithm called Deep Quality-Value (DQV) Learning. DQV uses temporal-difference learning to train a Value neural network and uses this network for training a second Quality-value network that learns to estimate state-action values. We first test DQV's update rules with Multilayer Perceptrons as function approximators on two classic RL problems, and then extend DQV with the use of Deep Convolutional Neural Networks, `Experience Replay' and `Target Neural Networks' for tackling four games of the Atari Arcade Learning environment. Our results show that DQV learns significantly faster and better than Deep Q-Learning and Double Deep Q-Learning, suggesting that our algorithm can potentially be a better performing synchronous temporal difference algorithm than what is currently present in DRL.
\end{abstract} 

\section{Introduction}
\label{sec:introduction} 

In Reinforcement Learning (RL), Temporal-Difference (TD) learning has become
a design choice which is shared among the most successful algorithms that are present in the field \cite{amiranashvili2018td}. Whether it is used in a Tabular-RL setting \cite{watkins1992q,hasselt2010double}, or in combination with a function approximator \cite{tsitsiklis1997analysis,sutton2009fast}, TD methods aim to learn a Value function, $V$, by directly bootstrapping their own experiences at different time-steps $t$. This is done with respect to a discount factor, $\gamma$, and a reward, $r$, which allow the computation of the TD-errors, $r_{t} + \gamma V(s_{t+1}) - V(s)$. Since the TD-errors can be computed directly while interacting with the environment, TD learning algorithms turn out to be a much faster and better performing alternative, when compared to other RL approaches such as Monte Carlo (MC) learning. In fact, the latter methods require the RL episodes to end before being able to estimate how to predict future rewards. 

The recent successful marriage between the most popular RL algorithms, and the use of Deep Neural Networks \cite{lecun2015deep}, has highlighted the power of TD-learning methods. Despite having to deal with the instabilities of a non-linear function approximator, such algorithms have led to remarkable results, which have made Deep Reinforcement Learning (DRL) one of the most popular fields within Machine Learning. Temporal-difference learning algorithms have been shown to be very efficient, both when applied on learning an action-value function \cite{mnih2015human,van2016deep,zhao2016deep,gu2016continuous,pong2018temporal}, and when used in combination with policy gradient methods \cite{castro2009temporal,lillicrap2015continuous,mnih2016asynchronous}.  

In this work we exploit the powerful capabilities of TD learning by extending a Tabular-RL algorithm, called QV$(\lambda)$ learning \cite{wiering2005qv}, so that its update rules can be successfully used in combination with Deep Artificial Neural Networks. Furthermore, we take advantage of the most popular strategies which are known to ensure stability when training DRL algorithms, in order to construct our novel DRL algorithm called Deep Quality-Value learning. 

The structure of this paper is as follows: in Section \ref{sec: preliminaries} we present the most important theoretical concepts of the field of RL, which serve as preliminary knowledge for our novel algorithm presented in Section \ref{sec: dqv_explanation}. We then present the methodological details about the experimental setup of this work in Section \ref{sec: methods} and the results that have been obtained in Section \ref{sec: results}. An analysis of the performances of our algorithm, together with a discussion about the relevance of our work, concludes the paper in Section \ref{sec: conclusion}.

\section{Preliminaries}
\label{sec: preliminaries}

Reinforcement Learning (RL) is the branch of Machine Learning in which artificial agents have to learn an optimal behavior while interacting with a particular environment \cite{sutton1998introduction}. An RL problem can be formalized as a Markov Decision Process (MDP) consisting of a set of possible states, $\cal S$, and a set of possible actions $\cal A$. By choosing an action $a \in \cal A$, the agent performs a state transition from state $s_t$ at time $t$ to $s_{t+1}$ that is defined by a transition probability distribution $p(s_{t+1} | s_{t}, a_{t})$. Associated to this transition probability there is an immediate reward function, $\Re (s_{t}, a_{t}, s_{t+1})$, that specifies the reward $r_t$ of being in state $s_{t+1}$ based on the action $a_t$ that the agent has taken in state $s_t$. The agent chooses which actions to perform based on its policy, which maps every state to a particular action $\pi:s \rightarrow a$. If a policy is followed, it is possible to compute for each state its Value ($V$):

\begin{equation} 
V^{\pi}(s) = \mathds{E} \left[\sum_{k=0}^{\infty} \gamma^{k} r_{t+k} \middle| s_t = s \right],
\label{eq: value_function}
\end{equation}

which corresponds to the expected cumulative reward that an agent will gain while being in state $s$, and by following policy $\pi$. The discount factor $\gamma$ is set between $[0,1]$, when its value is close to $0$ the agent will only consider the rewards that are obtained in an immediate future, while if its value is close to $1$, the agent will also aim to maximize the rewards obtained in a more distant future. Training an RL agent consists in finding an optimal policy, $\pi^{*}$, which maximizes the expected future reward from all possible states such that 

\begin{equation}
\pi^{*}(s)= \underset{\pi}{argmax} \; V^{\pi}(s).
\end{equation}

In addition to the $V$ function, there is a second function, $Q^{\pi}(s,a)$, which indicates how good or bad it is to select a particular action $a$ in state $s$ while following policy $\pi$. The goal is to find a policy that maximizes $Q$, which is known to correspond to satisfying the Bellman equation:   

\begin{align}
Q^{\pi}(s_{t}, a_{t}) = \sum_{s_{t+1}\in \cal S} p(s_{t+1} | s_{t}, a_{t}) 
						\bigg(\Re(s_{t}, a_{t}, s_{t+1}) +  \gamma \max_{a_{t+1} \in \cal A} Q^{\pi}
                        (s_{t+1}, a_{t+1})\bigg).
\end{align}

Once $Q$ is learned, the optimal policy can be easily found by choosing the action that maximizes $Q$ in each state. 

Whether the goal is to learn the $V$ or the $Q$ function, Tabular based RL aims to formulate the different RL update rules similar to Dynamic Programming algorithms, which makes it possible to exploit the Markov properties of such rules. However, it is well known that this approach is computationally very demanding and is only suitable for problems which have a limited state-action space. In order to deal with more complicated RL problems non parametric function approximators can be used \cite{busoniu2010reinforcement}. While several types of regression algorithms can be used to do so, such as Support Vector Machines \cite{martin2002line} or Tree-based approaches \cite{ernst2005tree}, a large popularity has been gained with the use of Deep Artificial Neural Networks. The combination of RL and Deep Learning has recently made it possible to master RL problems of the most various domains, ranging from videogames \cite{mnih2015human,wang2015dueling,mnih2016asynchronous,foerster2017stabilising,lample2017playing}, to boardgames \cite{lai2015giraffe,silver2016mastering,silver2017mastering} and complex robotics control problems \cite{abbeel2007application,lillicrap2015continuous,gu2017deep}.

\section{Deep Quality-Value (DQV) Learning}
\label{sec: dqv_explanation}

We now introduce our novel DRL algorithm called Deep Quality-Value (DQV) learning. We first present the online RL algorithm QV($\lambda$) together with its update rules, and then extend these update rules to objective functions that can be used to train Artificial Neural Networks (ANNs) for solving complex RL problems.  

\subsection{QV$(\lambda)$ Learning} 
 
QV$(\lambda)$ has been presented in \cite{wiering2005qv} and essentially consists of a value function RL algorithm that keeps track of both the $Q$ function and the $V$ function. Its main idea consists of learning the $V$ function through the use of temporal difference TD($\lambda$) learning \cite{sutton1988learning}, and using these estimates to learn the $Q$ function with an update rule similar to one step Q-Learning \cite{watkins1992q}. The main benefit of this approach is that the $V$ function might converge faster than the $Q$ function, since it is not directly dependent on the actions that are taken by the RL agent. After a transition,$\langle$ $s_{t}$, $a_{t}$, $r_{t}$, $s_{t+1}$ $\rangle$, QV$(\lambda)$ uses the TD$(\lambda)$ learning rule \cite{sutton1988learning} to update the $V$ function for all states:

\begin{equation}
V(s):= V(s) + \alpha \big[ r_{t} + \gamma V(s_{t+1}) - V(s_t) \big] e_{t}(s),
\label{eq: V_update}
\end{equation}

where $\alpha$ stands for the learning rate and $\gamma$ is again the discount factor. It is worth noting how the $V$ function is updated according to a similar learning rule as standard Q-Learning \cite{watkins1992q} to update the $Q$ function:
 
\begin{align}
Q(s_{t}, a_{t}) := Q(s_{t}, a_{t}) + \alpha \big[& r_{t} + \gamma \max_{a_{t+1} \in \cal A} Q(s_{t+1}, a_{t+1}) - Q(s_{t}, a_{t})\big].
\label{eq: Q_learning}
\end{align}

Besides this TD update, QV$(\lambda)$ also makes use of eligibility traces, defined as $e_t(s)$, that are necessary to keep track if a particular state has occurred before a certain time-step or not. These are updated for all states as follows:  

\begin{equation} 
e_{t}(s) = \gamma \lambda e_{t-1}(s) + \eta_t(s),
\end{equation}

where $\eta_t(s)$ is an indicator function that returns a value of $1$ whether a particular state occurred at time $t$ and $0$ otherwise. Before updating the $V$ function, QV$(\lambda)$ updates the $Q$ function first, and does this via the following update rule:

\begin{equation}
Q(s_{t}, a_{t}):= Q(s_{t}, a_{t}) + \alpha \big[r_{t} + \gamma V(s_{t+1}) - Q(s_{t}, a_{t}) \big].
\label{eq: Q_update}
\end{equation}

In \cite{wiering2005qv} it is shown that QV$(\lambda)$ outperforms different offline and online RL algorithms in Sutton's Dyna maze environment. However, this algorithm has so far only been used with tabular representations and Shallow Neural Networks \cite{wiering2009qv} and never in combination with Deep Artificial Neural Networks. Thus we now present its extension: Deep QV Learning. We show how to transform the update rules \ref{eq: V_update} and \ref{eq: Q_update} as objective functions to train ANNs, and how to make the training procedure stable with the use of `Experience Replay' \cite{lin1992self} and `Target Networks' \cite{mnih2015human}.

\subsection{DQV Learning}
\label{sec:dqv}

Since the aim is to approximate both the $V$ and the $Q$ function, we train two ANNs with two distinct objective functions. We thus define the parametrized $V$ neural network as $\Phi$ and the $Q$ neural network as $\theta$. In order to build the two objective functions it is possible to simply express QV-Learning's update rules in Mean Squared Error terms similarly to how DQN addresses the Q-Learning update rule \ref{eq: Q_learning}:

\begin{equation}
L_{\theta}=\mathds{E}\big[(r_{t}+\gamma \max_{a_{t+1}\in \cal A} Q(s_{t+1}, a_{t+1}, \theta)-Q(s_{t},a_{t},\theta))^{2}\big].
\label{eq: DQN}
\end{equation}

Hence we obtain the following objective function when aiming to train the $V$ function:   

\begin{equation}
L_{\Phi} = \mathds{E} \big[(r_{t} + \gamma V(s_{t+1}, \Phi) - V(s_{t}, \Phi))^{2}\big],
\label{eq: v_update_ann}
\end{equation}

while the following one can be used to train the $Q$ function:

\begin{equation}
L_{\theta} = \mathds{E} \big[(r_{t} + \gamma V(s_{t+1}, \Phi) - Q(s_{t}, a_{t}, \theta))^{2}\big].
\label{eq: q_update_ann}
\end{equation}

By expressing the update rules \ref{eq: V_update} and \ref{eq: Q_update} as such it becomes possible to minimize the just presented objective functions by gradient descent.

QV-Learning is technically an online reinforcement learning algorithm since it assumes that its update rules get computed each time an agent has performed an action and has observed its relative reward. However, when it comes to more complex control problems (like the games of the Arcade Learning Environment, ALE, presented in \cite{mnih2015human}), training a Deep Convolutional Neural Network (DCNN) in an online setting is computationally not suitable. This would in fact make each experience usable for training exactly one time, and as a result, a very large set of experiences will have to be collected to properly train the ANNs. To overcome this issue it is possible to make use of `Experience Replay' \cite{mnih2015human,lillicrap2015continuous,zhao2016deep}, a technique which allows to learn from past episodes multiple times and that has proven to be extremely beneficial when tackling RL problems with DCNNs.

\textbf{Experience Replay} essentially consists of a memory buffer, $D$, of size $N$, in which experiences are stored in the form $\langle$ $s_{t}$, $a_{t}$, $r_{t}$, $s_{t+1}$ $\rangle$. Once this memory buffer is filled with a large set of these quadruples, $\mathscr{N}$, it becomes possible to randomly sample batches of such experiences for training the model. By doing so the RL agent can learn from previous experiences multiple times and does not have to rely only on the current $\langle$ $s_{t}$, $a_{t}$, $r_{t}$, $s_{t+1}$ $\rangle$ quadruple for training. In our experiments we use an Experience Replay buffer that stores the most recent 400,000 frames that come from the ALE, we uniformly sample batches of $32$ experiences $\langle$ $s_{t}$, $a_{t}$, $r_{t}$, $s_{t+1}$ $\rangle \sim U(D)$, and use them for optimizing the loss functions \ref{eq: v_update_ann} and \ref{eq: q_update_ann} each time the agent has performed an action. Training starts as soon as 50,000 frames have been collected in the buffer.

Besides making it possible to exploit past experiences multiple times, training from Experience Replay is also known for improving the stability of the training procedure. In fact, since the trajectories that are used for optimizing the networks get randomly sampled from the memory buffer, this makes the samples used for training much less correlated to each other, which yields more robust training. A second idea that serves the same purpose when it comes to TD algorithms has been proposed in \cite{mnih2015human}, and is known as the `Target Neural Network'. 

\textbf{Target Neural Network:} it consists of a separate ANN that is specifically designed for estimating the targets ($y_{t}$) that are necessary for computing the TD errors. This ANN has the exact same structure as the one which is getting optimized, but its weights do not change each time RL experiences are sampled from the Experience Replay buffer to train the value function. On the contrary, the weights of the target network are temporally frozen, and only periodically get updated with the parameters of the main network. In the classic DQN scenario the target network is parametrized as $\theta^{-}$, while the Deep Q-Network uses $\theta$, leading to a slight modification of the loss function presented in Equation \ref{eq: DQN}:
 
\begin{equation}
L_{\theta}=\mathds{E}\big[(r_{t}+\gamma \max_{a_{t+1}\in \cal A} Q(s_{t+1}, a_{t+1}, \theta^{-})-Q(s_t,a_t,\theta))^{2}\big].
\end{equation}

Since DQV also computes TD errors we include a specifically designed Value-Target-Network to our algorithm, which, similarly to what happens in DQN, slightly modifies the original loss functions presented in Equations \ref{eq: v_update_ann} and \ref{eq: q_update_ann} to:

\begin{equation}
L_{\Phi} = \mathds{E} \big[(r_{t} + \gamma V(s_{t+1}, \Phi^{-}) - V(s_{t}, \Phi))^{2}\big]
\end{equation}

and 

\begin{equation}
L_{\theta} = \mathds{E} \big[(r_{t} + \gamma V(s_{t+1}, \Phi^{-}) - Q(s_{t}, a_{t}, \theta))^{2}\big].
\end{equation}

For our experiments we update the Value-Target-Network $\Phi^{-}$ with the weights of our original Value Network $\Phi$ every 10,000 actions as defined by the hyperparameter $c$.

We have now defined all elements of our novel DRL algorithm which is summarized by the pseudocode presented in Algorithm \ref{alg: dqv_algorithm_full}.

\begin{algorithm}[ht!]
\caption{DQV Learning}\label{alg: dqv_algorithm_full}
\begin{algorithmic}[1]
\Require{Experience Replay Stack $D$ with size $N$}{}
\Require{$Q$ network with parameters $\theta$}{}
\Require{$V$ networks with parameters $\Phi$ and $\Phi^{-}$} {}
\Require{total\_a = 0}
\Require{total\_e = 0}
\Require{c = $10000$}
\Require{$\mathscr{N} = 50000$}

\For{$e$ $\in$ EPISODES}
\While{$e$ $\neg$ over}
	\State{observe $s_{t}$}
    \State{select $a_t$ $\in$ $\cal A$ for $s_{t}$ with policy $\pi$}
    \State{get $r_{t}$ and $s_{t+1}$}
    \State{store $e$ as $\langle$ $s_{t}$, $a_{t}$, $r_{t}$, $s_{t+1}$ $\rangle$ in $D$}
    \State{total\_e += 1}
    \If{total\_e $\geqq \mathscr{N}$}
    	\State{sample minibatch of $32$ $e$ from $D$}
        \If{$s_{t+1}$ $\in$ $e$ is over}
        	\State{$y_{t} := r_{t}$}	
        \Else
			\State{$y_{t} := r_{t} + \gamma V(s_{t+1}, \Phi^{-})$}
         \EndIf
     	\State{$\theta := \underset{\theta}{\argmin} \mathds{E}[(y_{t} - Q(s_{t}, a_{t}, \theta))^{2}]$}
        \State{$\Phi := \underset{\Phi}{\argmin} \mathds{E}[(y_{t} - V(s_{t}, \Phi))^{2}]$}
        \State{total\_a += 1}
		\If{total\_a = $c$}
        	\State{$\Phi^{-}$ := $\Phi$}
            \State{total\_a := 0}
         \EndIf
    \EndIf
\EndWhile
\EndFor

\end{algorithmic}
\end{algorithm}

\section{Experimental Setup}
\label{sec: methods}
 
\begin{figure}[ht!]
\begin{minipage}{.5\textwidth}
  \includegraphics[width=1\linewidth]{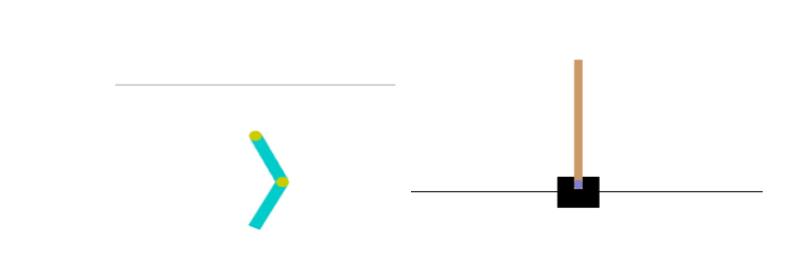}
  \end{minipage}%
\begin{minipage}{.5\textwidth}
  \includegraphics[width=1\linewidth]{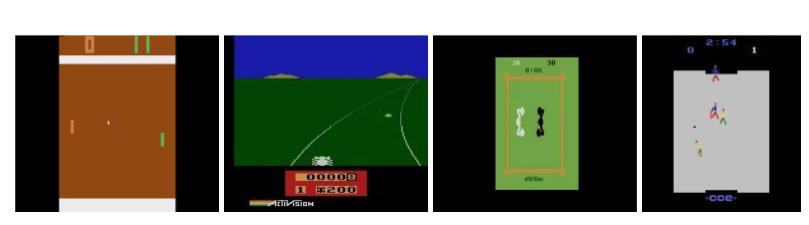}
  \end{minipage}
  \caption{A visualization of the six RL environments that have been used in our experiments as provided by the Open-AI Gym package \cite{brockman2016openai}. From left to right: \textit{Acrobot, Cartpole, Pong, Enduro, Boxing, Ice-Hockey.}}
\label{fig: atari_environments}
\end{figure}

For our experiments we use two groups of environments that are both provided by the Open-AI Gym package \cite{brockman2016openai}. The first group is intended to explore DQV's performance on a set of RL tasks that do not rely on computationally expensive GPU support and make it possible to quickly fine tune all the necessary hyperparameters of the algorithm. It consists of two environments that simulate two classic control problems that are well known in RL literature: \textit{Acrobot} \cite{sutton1996generalization} and \textit{Cartpole} \cite{barto1983neuronlike}. The second group of environments is more challenging and consists of several Atari 2600 games. In particular we investigate DQV's performances on the games \textit{Pong}, \textit{Enduro}, \textit{Boxing} and \textit{Ice-Hockey}. A visualization of such environments can be seen in Figure \ref{fig: atari_environments}. In all the experiments we compare the performance of DQV with two other well known self implemented TD DRL algorithms: DQN \cite{mnih2015human} and DDQN \cite{van2016deep}. 

We use a two hidden layer Multilayer Perceptron (MLP) activated by a ReLU non linearity ($f(x) = max (0,x)$) when we test the DRL algorithms on the first two environments. While a three hidden layer DCNN, followed by a fully connected layer as originally presented in \cite{mnih2015human}, when tackling the Atari $2600$ games. For the latter set of environments we use the \verb#Deterministic-v4# versions of the games, which, as proposed in \cite{mnih2015human}, use `Frame-Skipping', a design choice which lets the agent select an action every $4$th frame instead of every single one. Furthermore, we use the standard Atari-preprocessing scheme in order to resize each frame to an $84 \times 84$ gray-scaled matrix.

The Adam optimizer \cite{kingma2014adam} has been used for optimizing the MLPs while the RMSprop optimizer \cite{tieleman2012lecture} has been used for the DCNNs. As DQV's exploration strategy we use the well known epsilon-greedy approach with an initial $\epsilon$ value set to $0.5$ which gets linearly decreased to $0.1$. The discount factor $\gamma$ is set to $0.99$. Lastly, since the scale of the rewards changes from game to game, we `clip' all the rewards to the $[-1,1]$ range. 

\section{Results}
\label{sec: results}

We first evaluate the performance that DQV has obtained on the \textit{Acrobot} and \textit{Cartpole} environments where we have used MLPs as function approximators. Given the relative simplicity of these problems we did not integrate DQV with the Value-Target-Network. In fact, we have only optimized the two different MLPs according to the objective functions \ref{eq: v_update_ann} and \ref{eq: q_update_ann} presented in Section \ref{sec:dqv}. Regarding the Experience Replay buffer we did not include it in the \textit{Acrobot} environment while it is used in the \textit{Cartpole} one.
However its size is far smaller than the one which has been presented in Algorithm \ref{alg: dqv_algorithm_full}, in fact, since the problem tackled is much simpler, we only store $200$ experiences which get randomly sampled with a batch size of $16$.

We then present the results obtained on the Atari $2600$ games, where instead of MLPs, we have used DCNNs and have integrated the algorithm with a larger Experience Replay buffer and the Value-Target-Network. The algorithm used is thus the one presented in Algorithm \ref{alg: dqv_algorithm_full}. We report on the $x$ axis of all figures the amount of RL episodes that have been used to train the different algorithms, and on the $y$ axis the cumulative reward that has been obtained by the agents in each episode. All results report the average cumulative reward which has been obtained over $5$ different simulations with $5$ different random seeds. Please note that we smoothed the reward function with a `Savgol Filter' to improve the interpretability of the plots. 

\begin{figure}
\begin{minipage}{.5\textwidth}
  \includegraphics[width=1\linewidth]{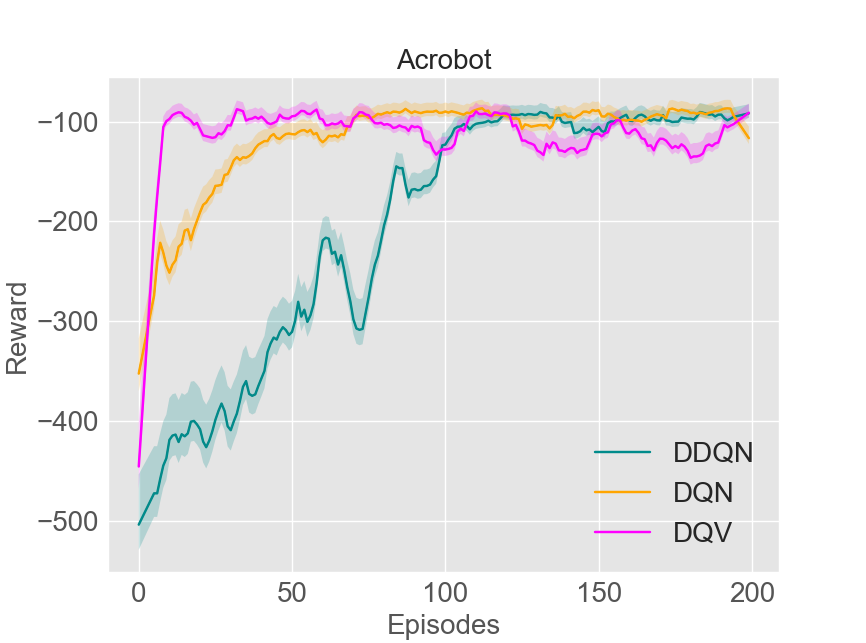}
  \label{fig:test1}
\end{minipage}%
\begin{minipage}{.5\textwidth}
  \includegraphics[width=1\linewidth]{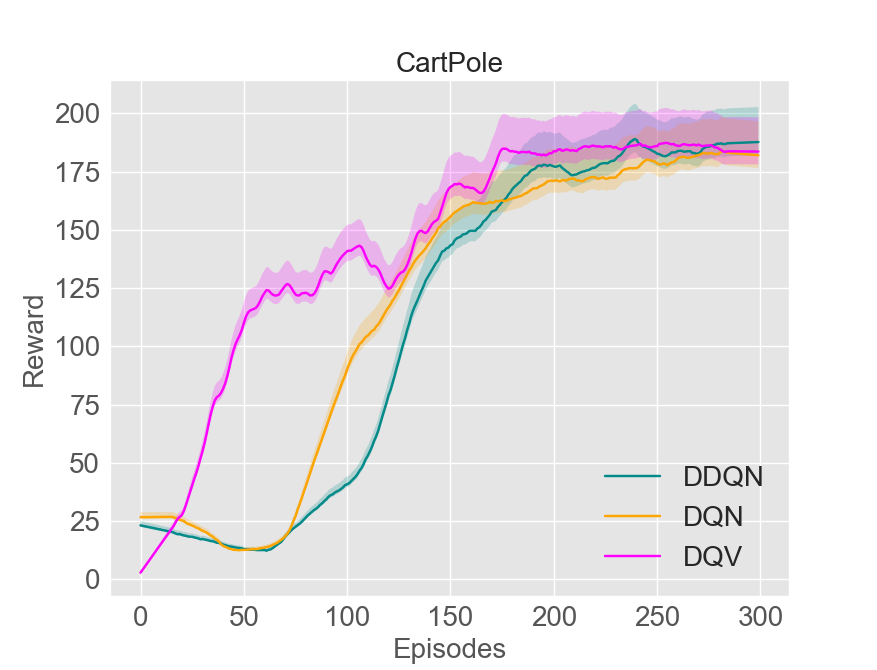}
  \label{fig:test2}
\end{minipage}
\caption{The results that have been obtained when using DQV in combination with an MLP on two classic RL problems (Acrobot and Cartpole). DQV learns significantly faster when compared with DQN and DDQN.}
\label{fig: dqv_mlp}
\end{figure}

\begin{figure}
\begin{minipage}{.5\textwidth}
  \includegraphics[width=1\linewidth]{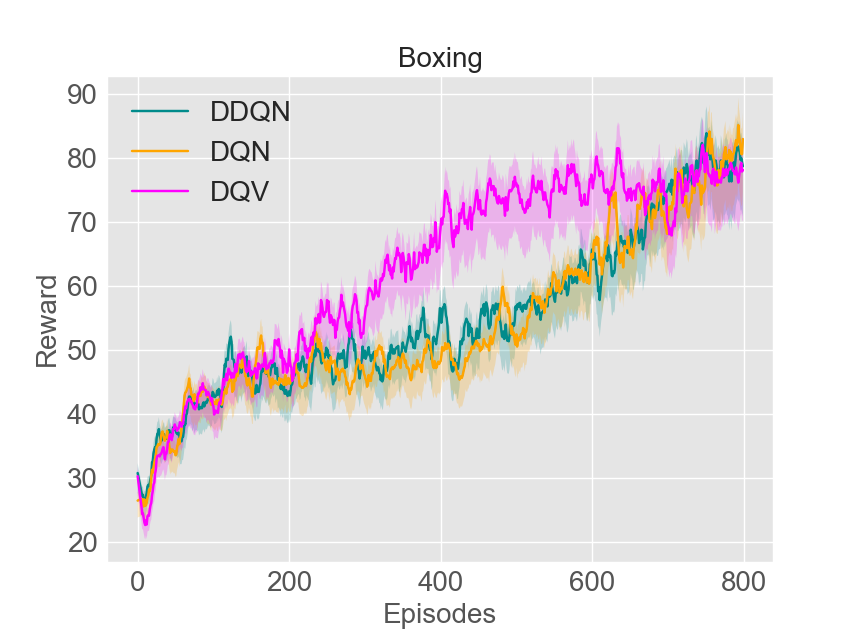}
  \label{fig:test2}
\end{minipage}%
\begin{minipage}{.5\textwidth}
  \includegraphics[width=1\linewidth]{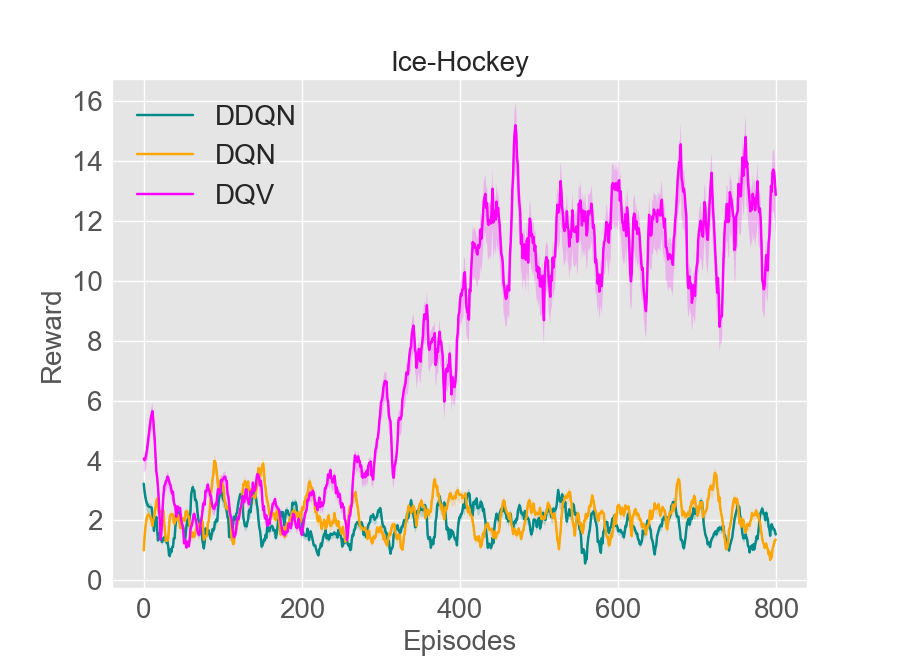}
  \label{fig:test2}
\end{minipage}
\caption{The results which have been obtained when using DQV in combination with Deep Convolutional Neural Networks, Experience Replay and Target Neural Networks on four Atari games. It can be seen that DQV learns significantly faster on the games \textit{Pong} and \textit{Boxing} while it also yields better results on the games \textit{Enduro} and \textit{Ice-Hockey}.}
\label{fig:dqv_dcnns}
\end{figure}

The results obtained on \textit{Acrobot} and \textit{Cartpole} show how promising the update rules \ref{eq: v_update_ann} and \ref{eq: q_update_ann} can be when used in combination with an MLP. In fact as shown in Figure \ref{fig: dqv_mlp} it is possible to see that DQV is able to significantly outperform DQN and DDQN in both environments. It is however worth noting that the problems which have been tackled in this set of experiments are not particularly complex. Hence, in order to concretely establish the performance of our novel algorithm, the Atari $2600$ games serve as an ideal testbed. The results presented in Figure \ref{fig:dqv_dcnns} show how DQV is again able to systematically outperform DQN and DDQN in all the experiments we have performed. When DQV is tested on the games \textit{Pong} and \textit{Boxing} it can be seen that it learns significantly faster when compared to the other two algorithms. This is particularly true for the game \textit{Pong} which gets solved in less than $400$ episodes, making DQV more than two times faster than DQN and DDQN. Similar results have been obtained on the \textit{Boxing} environment, where DQV approaches the maximum possible cumulative reward in the game ($\approx 80$) almost $300$ episodes before the other two algorithms. On the \textit{Enduro} and \textit{Ice-Hockey} games we can see that DQV does not only learn faster, but also obtains much higher cumulative rewards during the game. These rewards are almost twice as high in the \textit{Enduro} environment, while the results also show that DQV is the only DRL algorithm which is able to optimize the policy over $800$ episodes when it is tested on the \textit{Ice-Hockey} game. The latter results are particularly interesting since they correspond to the first and only case so far, in which we have observed that DQN and DDQN are not able to improve the policy during the time that is required by DQV to do so.

\begin{figure}
\begin{minipage}{.5\textwidth}
  \includegraphics[width=1\linewidth]{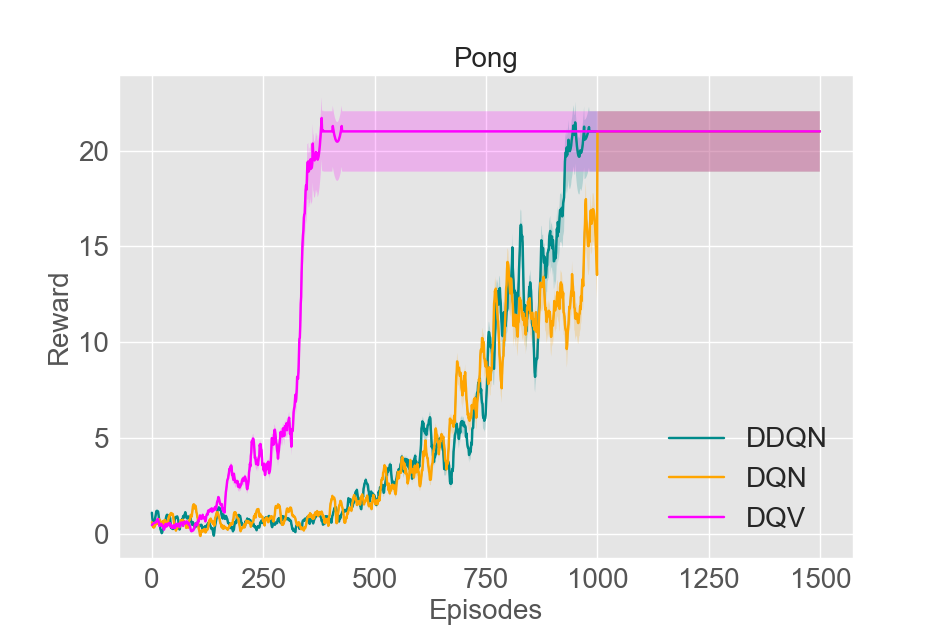}
  \label{fig:test1}
\end{minipage}%
\begin{minipage}{.5\textwidth}
  \includegraphics[width=1\linewidth]{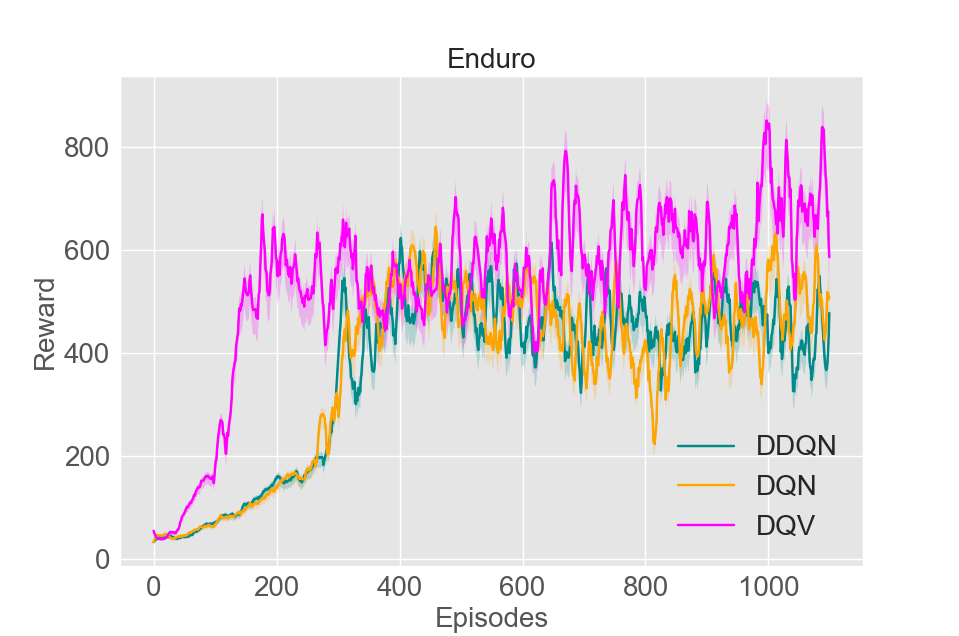}
  \label{fig:test2}
\end{minipage}
\end{figure}

\section{Discussion and Conclusion}
\label{sec: conclusion}

Besides introducing a novel DRL algorithm, we believe that the methodologies and results proposed in this paper contribute to the field of DRL in several important ways. We showed how it is possible to build upon existing Tabular-RL work, in order to successfully extend the set of DRL algorithms which are currently present in the field. So far, the results obtained by DQV-Learning are very promising and definitely suggest that our algorithm is a better alternative if compared to the well known DQN and DDQN algorithms. Furthermore, DQV also reaffirms the benefits of TD-Learning. More specifically, it is particularly interesting to see how effective the use of the $V$ neural networks is, when learning an action-value function with the $Q$ network. When QV($\lambda$) has been initially introduced, one of the main intuitions behind the algorithm relied on the fact that the $V$ function could converge faster than the $Q$ function, which can make the algorithm learn faster. The results obtained in this paper support this idea, with DQV being the fastest algorithm in all the experiments that we have performed. 

Moreover, we also noticed how important it is to integrate a DRL algorithm with a specifically designed Target-Network that is explicitly built for estimating the target values while learning. It is worth noting however, that unlike DQN and DDQN, this Target-Network is required for the $V$ function and not the $Q$ function, because the $V$ function is used to update both value functions. Lastly, we also remark the importance of including an Experience-Replay buffer in the algorithm when tackling complex RL problems as the ones provided by the Arcade Learning Environment.
We aim to research more in detail the role of the Target-Network. In fact, there is to the best of our knowledge no real formal theory that motivates the need for this additional neural architecture. Nevertheless, it is a design choice which is well known to be useful in the field. With DQV there is a new algorithm that makes use of it, and which could help for gaining a better understanding of this additional neural architecture.  

The main strength of DQV certainly relies on the speed that our algorithm obtains for learning. However, we are aware that more sophisticated approaches such as the Asynchronous-Advantage-Actor-Critic (A3C) algorithm have already been proposed in the field for dealing with long RL training times. As future work we want to work on a multi-threaded version of DQV, and investigate if, in case its update rules are used for training multiple agents in parallel, our algorithm could perform better and faster than A3C. Furthermore, by estimating both the $V$ function and the $Q$ function, DQV-Learning allows for a novel method to estimate the Advantage function, and we want to study if this can help algorithms such as A3C to perform better. Finally, we want to take into account the several contributions that have been proposed in the literature for making DQN more data efficient and faster \cite{hessel2017rainbow}. We will perform a similar ablation study on DQV, and investigate whether such improvements will be as successful when applied to our algorithm and result in state-of-the-art performance on many different Atari 2600 games.

\subsection*{Acknowledgements}
Matthia Sabatelli acknowledges the financial support of BELSPO, Federal Public Planning Service Science Policy, Belgium, in the context of the BRAIN-be project.

\bibliographystyle{plain}
\bibliography{example.bib}

\end{document}